\documentclass{article} 
\usepackage{nips12submit_e,times}
\usepackage{bm}
\usepackage{multirow}
\newcommand*\N{\mathcal{N}}
\newcommand*\Vec[1]{\bm{{#1}}}

\usepackage{graphicx}
\usepackage{booktabs}
\usepackage{subfig}
\usepackage{ed}
\usepackage{xfrac}

\title{Unsupervised Feature Learning for low-level Local Image Descriptors}

\author{
Christian Osendorfer, Justin Bayer, Sebastian Urban, Patrick van der Smagt\\
Technische Universit\"{a}t M\"{u}nchen\\
\texttt{\{osendorf, bayerj, surban, smagt\}@in.tum.de} \\
}

\nipsfinalcopy 

\begin{document}

\maketitle

\begin{abstract}
Unsupervised feature learning has shown impressive results for a wide range of input modalities, in particular for object classification tasks in computer vision. Using a large amount of unlabeled data, unsupervised feature learning methods are utilized to construct high-level representations that are discriminative enough for subsequently trained supervised classification algorithms. However, it has never been \emph{quantitatively} investigated yet how well unsupervised learning methods can find \emph{low-level representations} for image patches without any additional supervision. In this paper we examine the performance of pure unsupervised methods on a low-level correspondence task, a problem that is central to many Computer Vision applications. We find that a special type of Restricted Boltzmann Machines (RBMs) performs comparably to hand-crafted descriptors. Additionally, a simple binarization scheme produces compact representations that perform better than several state-of-the-art descriptors.
%
\end{abstract}

\section{Introduction}
In this paper we tackle a recent computer vision dataset \cite{brown2010a} from the viewpoint of unsupervised feature learning. Why yet another dataset? There are already enough datasets that serve well for evaluating feature learning algorithms. In particular for feature learning from image data, several well-established benchmarks exist: Caltech-101 \cite{fei-fei2007a}, CIFAR-10 \cite{krizhevsky2009a}, NORB \cite{lecun2004a}, to name only a few. Notably, these benchmarks are all object classification tasks. Unsupervised learning algorithms are evaluated by considering how well a subsequent supervised classification algorithm performs on high-level features that are found by aggregating the learned low-level representations \cite{coates2011b}. We think that mingling these steps makes it difficult to assess the quality of the unsupervised algorithms. A more direct way is needed to evaluate these methods, preferably where a subsequent supervised learning step is completely optional. 

We are not only at odds with the methodology of evaluating unsupervised learning algorithms. General object classification tasks are always based on orientation- and scale-rectified pictures with objects or themes firmly centered in the middle. We are looking for a dataset where it is possible to show that unsupervised feature learning is beneficial to the wide range of Computer Vision tasks beyond object classification, like tracking, stereo vision, panoramic stitching or structure from motion.
%
%
%
%
%
%
%
One might argue, that object classification acts as a good proxy for all these other tasks but this hypothesis has not shown to be correct either theoretically or through empirical evidence. Instead, we chose the most general and direct task that can be be applied to \emph{low-level representations}: matching these representations, i.e.\ determining if two data samples are similar given their learned representation.

Matching image descriptors is a central problem in Computer Vision, so hand-crafted descriptors are always evaluated with respect to this task \cite{mikolajczyk2005a}. Given a dataset of labeled correspondences, \emph{supervised} learning approaches will find representations \emph{and} the accompanying distance metric that are optimized with respect to the induced similarity measure. It is remarkable that hand-engineered descriptors perform well under this task \emph{without the need to learn such a measure} for their representations in a supervised manner. 

To the best of our knowledge it has never been investigated whether any of the many unsupervised learning algorithms developed over the last couple of years can match this performance without relying on any supervision signals. While we propose an additional benchmark for unsupervised learning algorithms, we do not introduce a new learning algorithm. We rather investigate the performance of the Gaussian RBM (GRBM) \cite{tang2012a}, its sparse variant (spGRBM) 
\cite{nair2009a} and the mean covariance RBM (mcRBM) \cite{ranzato2010b} without any supervised learning with respect to the matching task. As it turns out, the mcRBM performs comparably to hand-engineered feature descriptors. In fact using a simple heuristic, the mcRBM produces a \emph{compact binary} descriptor that performs better than several state-of-the-art hand-crafted descriptors.

We begin with a brief description of the dataset used for evaluating the matching task, followed by a section on details of the training procedure. In section~\ref{sec:res} we present our results, both quantitatively and qualitatively and also mention other models that were tested but not further analyzed because of overall bad performance. Section~\ref{sec:conc} concludes with a brief summary and an outlook for future work. A review of GRBMs, spGRBMs and mcRBMs is provided in the appendix, section~\ref{sec:mod}, for completeness.  

\paragraph{Related work}
Most similar in spirit to our work are \cite{coates2012a, krizhevsky2011a, le2012a}: Like us, \cite{coates2012a, le2012a} are interested in the behavior of unsupervised learning approaches without any supervised steps afterwards. Whereas both investigate high-level representations. \cite{krizhevsky2011a} learns a compact, binary representation with a very deep autoencoder in order to do fast content-based image search (\emph{semantic hashing}, \cite{salakhutdinov2008a}). Again, these representations are studied with respect to their capabilities to model high-level object concepts. Additionally, various algorithms to learn high-level correspondences have been studied \cite{chopra2005a, susskind2011a, huang2012a} in recent years.

Finding (compact) low-level image descriptors should be an excellent machine learning task: Even hand-designed descriptors have many free parameters that cannot (or should not) be optimized manually. Given ground truth data for correspondences, the performance of supervised learning algorithms is impressive \cite{brown2010a}. Very recently, boosted learning with image gradient-based weak learners has shown excellent results \cite{trzcinski2012a, trzcinski2012c} on the same dataset used in this paper. See section 2 of \cite{trzcinski2012a} for more related work in the space of supervised metric learning.

\section{Dataset}\label{sec:data}
\begin{figure}[tp]
\begin{center}
\includegraphics[width=\linewidth]{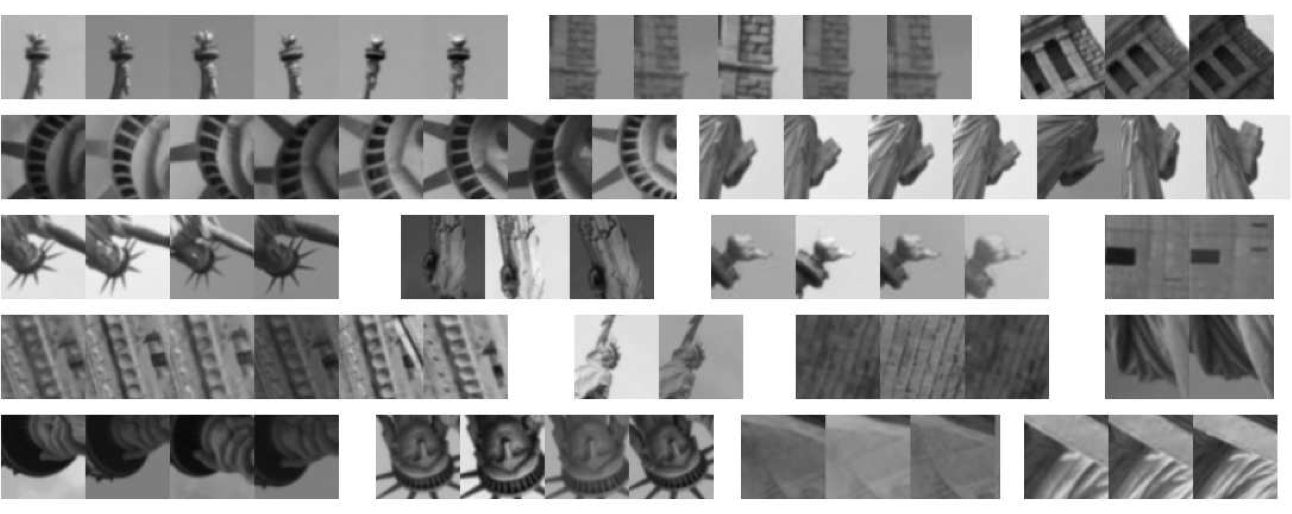}
\end{center}
\caption{Patch correspondences from the Liberty dataset. Note the wide variation in lighting, viewpoint and level of detail. The patches are centered on interest points but otherwise can be considered random, e.g.\ there is no reasonable notion of an object boundary possible. Figure taken from \cite{brown2010a}.}
\label{fig:patches}
\end{figure}

At the heart of this paper is a recently introduced dataset for discriminative learning of local image descriptors \cite{brown2010a}. It attempts to foster learning optimal low-level image representations using a large and realistic training set of patch correspondences. The dataset is based on more than 1.5 million image patches ($64 \times 64$ pixels) of three different scenes: the Statue of Liberty (about 450,000 patches), Notre Dame (about 450,000 patches) and Yosemite's Half Dome (about 650,000 patches). The patches are sampled around interest points detected by Difference of Gaussians \cite{lowe2004a} and are normalized with respect to scale and orientation\footnote{A similar dataset of patches centered on multi-scale Harris corners is also available.}. As  shown in Figure~\ref{fig:patches},  the dataset has a wide variation in lighting conditions, viewpoints, and scales. 

The dataset contains also approximately 2.5 million image correspondences. Correspondences between image patches are established via dense surface models obtained from stereo matching (stereo matching, with its epipolar and multi-view constraints, is a much easier problem than unconstrained 2D feature matching). The exact procedure to establish correspondences is more involved and described in detail in \cite[Section II]{brown2010a}. Because actual 3D correspondences  are used, the identified 2D patch correspondences show substantial perspective distortions resulting in a much more realistic dataset than previous approaches \cite{lepetit2006a,mikolajczyk2005a}. The dataset appears very similar to an earlier benchmark of the same authors \cite{winder2007a}, yet the correspondences in the novel dataset resemble a much harder problem. The error rate at 95\% detection of correct matches for the SIFT descriptor \cite{lowe2004a} raises from 6\% to 26\%, the error rate for evaluating patch similarity in pixel space (using normalized sum squared differences) raises from 20\% to at least 48\% (all numbers are take from \cite{winder2007a} and \cite{brown2010a} respectively), for example. In order to facilitate comparison of various descriptor algorithms a large set of predetermined match/non-match patch pairs is provided. For every scene, sets comprising between 500 and 500,000 pairs (with 50\% matching and 50\% non-matching pairs) are available. 

We don't argue that this dataset subsumes or substitutes any of the previously mentioned benchmarks. Instead, we think that it can serve to complement those. It constitutes an excellent testbed for unsupervised learning algorithms: Experiments considering self-taught learning \cite{raina2007a}, effects of semi-supervised learning, supervised transfer learning over input distributions with a varying degree of similarity (the scenes of Statue of Liberty and Notredame show architectural structures, while Half Dome resembles a typical natural scenery) and the effect of enhancing the dataset with arbitrary image patches around keypoints can all be conducted in a controlled environment. Furthermore, end-to-end trained systems for (large) classification problems (like \cite{krizhevsky2012a, ciresan2012a}) can be evaluated with respect to this type of data distribution and task.

\section{Training Setup}\label{sec:train}
Different to \cite{brown2010a}, our models are trained in an unsupervised fashion on the available patches. We train on one scene (400,000 randomly selected patches from this scene) and evaluate the performance on the test set of every scene. This allows us to investigate the self-taught learning paradigm \cite{raina2007a}. We also train on all three scenes jointly (represented by 1.2 million image patches) and then evaluate again every scene individually.

\subsection{GRBM/spGRBM}\label{subsec:train:grbm}
The GRBM and spGRBM (see Appendix, section~\ref{sec:spgrbm}) only differ in the setting of the sparsity penalty $\lambda_{\mathrm{sp}}$, all other settings are the same. We use $\mathrm{CD}_1$ \cite{hinton2002a} to compute the approximate gradient of the log-likelihood and the recently proposed rmsprop \cite{tieleman2012a} method as gradient ascent method. Compared to standard minibatch gradient ascent, we find that rmsprop is a more efficient method with respect to the training time necessary to learn good representations: it takes at most half of the training time necessary for standard minibatch gradient ascent.

Before learning the parameters, we first scale all image patches to $16\times 16$ pixels. Then we preprocess all training samples by subtracting the vectors' mean and dividing by the standard deviation of its elements. This is a common practice for visual data and corresponds to local brightness and contrast normalization. \cite[Section 2.2]{tang2012a} gives also a theoretical justification for why this preprocessing step is necessary to learn a reasonable precision matrix $\Vec{\Lambda}$. We find that this is the only preprocessing scheme that allows GRBM and spGRBM to achieve good results. In addition, it is important to learn $\Vec{\Lambda}$---setting it to the identity matrix, a common practice \cite{hinton2010b}, also produces dissatisfying error rates. Note that originally it was considered that learning $\Vec{\Lambda}$ is mostly important when one wants to find a good density (i.e.\ generative) model of the data. 

Both GRBM and spGRBM have 512 hidden units. The elements of $\Vec{W}$ are initialized according to $\N(0, 0.1)$, the biases are initialized to 0. rmsprop uses a learning rate of $0.001$, the decay factor is $0.9$, the minibatch size is 128. We train both models for 10 epochs (this takes about 15 minutes on a consumer GPU for 400000 patches). For the spGRBM we use a sparsity target of $\rho=0.05$ and a sparsity penalty of $\lambda_{\mathrm{sp}}=0.2$. spGRBM is very sensitive to settings of $\lambda_{\mathrm{sp}}$ \cite{swersky2012a}---setting it too high results in dead representations (samples that have no active hidden units) and the results deteriorate drastically.

\subsection{mcRBM}
\label{subsec:train:mcRBM}
mcRBM (see Appendix, section~\ref{sec:app:mcrbm}) training is performed using the code from \cite{ranzato2010b}. We resample the patches to $16 \times 16$ pixels. Then the samples are preprocessed by subtracting their mean (patchwise), followed by PCA whitening, which retains 99\% of the variance. The overall training procedure (with stochastic gradient descent) is identical to the one described in \cite[Section 4]{ranzato2010b}. We train all architectures for a total of 100 epochs, however updating $\Vec{P}$ is only started after epoch 50. We consider two different mcRBM architectures: The first has 256 mean units, 512 factors and 512 covariance units. $\Vec{P}$ is not constrained by any fixed topography. We denote this architecture by \emph{mcRBM}$(256,512/512)$. The second architecture is concerned with learning more compact representations: It has 64 mean units, 576 factors and 64 covariance units. $\Vec{P}$ is initialized with a two-dimensional topography that takes $5 \times 5$ neighborhoods of factors with a stride equal to 3. We denote this model by \emph{mcRBM}$(64,576/64)$. On a consumer grade GPU it takes 6 hours to train the first architecture on 400000 samples and 4 hours to train the second architecture on the same number of samples.

\section{Results}
\label{sec:res}
For the results presented in this section (Table~\ref{tab:res}) we follow the evaluation procedure of \cite{brown2010a}: For every scene (Liberty (denoted by LY), Notredame (ND) and Half Dome (HD)), we use the labeled dataset with 100,000 image pairs to assess the quality of a trained model on this scene. In order to save space we do not present ROC curves and only show the results in terms of the \emph{95\% error rate} which is the percent of incorrect matches when 95\% of the true matches are found: After computing the respective distances for all pairs in a test set, a threshold is determined such that 95\% of all matching pairs have a distance below this threshold. Non-matching pairs with a distance below this threshold are considered incorrect matches. 

\begin{table}[t]
    \centering
    \tabcolsep=0.11cm
    \small
    \subfloat[]{%
    \begin{tabular}{llccc}
        \toprule
        & & \multicolumn{3}{c}{\textbf{Test set}} \\
        \textbf{Method} & \textbf{Training set} & LY & ND & HD \\
        \toprule
        SIFT & -- & 28.1 & 20.9 & 24.7 \\

        \\
        & LY & 47.6 & 33.5 & 41.4 \\
        GRBM& ND & 50.0 & 33.4 & 42.5 \\
        $(L_1\ell_1)$& HD & 49.0 & 34.0 & 41.5 \\
        & LY/ND/HD & 48.7 & 33.5 & 42.1 \\

        \\
        & LY & 37.9 & 26.9 & 34.3  \\
        spGRBM & ND & 40.0 & 28.0 & 35.4  \\
        $(L_1\ell_1)$& HD & 39.1 & 27.9 & 34.9 \\
        & LY/ND/HD & 37.5 & 26.6 & 33.6 \\

        \\
        & LY & 31.3 & 25.1 & 34.5 \\
        mcRBM& ND & 34.0 & 25.6 & 33.0 \\
        $(L_1\ell_2)$& HD & 31.2&  22.3 & 25.7 \\
        & LY/ND/HD & 30.8 & 24.8 & 33.3 \\

        \\
        & LY & 34.7 & 24.2 & 38.6 \\
        mcRBM & ND & 33.3 & 24.8 & 44.9 \\
        (JSD) & HD & 29.9 & 22.7 & 37.6 \\
        & LY/ND/HD & 30.0 & 23.1 & 39.8 \\

        \bottomrule
    \end{tabular}}
    \qquad
    \subfloat[]{%
    \begin{tabular}{llccc}
        \toprule
        & & \multicolumn{3}{c}{\textbf{Test set}} \\
        \textbf{Method} & \textbf{Training set} & LY & ND & HD \\
        \toprule
        SIFT & -- & 31.7 & 22.8 & 25.6 \\
        BRIEF & -- & 59.1 & 54.5 & 54.9 \\
        BRISK & -- & 79.3 & 74.8 & 73.2 \\
        SURF & -- & 54.0 & 45.5 & 43.5\\

        \\
        & LY & -- & 16.9 & 22.8 \\[-0.9ex]
        \raisebox{1.1ex}{BinBoost}& ND & 20.4 & -- & 18.9 \\[-0.9ex]
        \raisebox{1.1ex}{(8 bytes)}& HD & 21.6 & 14.5 & -- \\

        \\
        & LY & -- & 31.1 & 34.4 \\[-0.9ex]
        \raisebox{1.1ex}{ITQ-SIFT}& ND & 37.0 & -- & 34.3 \\[-0.9ex]
        \raisebox{1.1ex}{(8 bytes)}& HD & 37.3 & 30.5 & -- \\

        \\
        & LY & -- & 43.1 & 47.2 \\[-0.9ex]
        \raisebox{1.1ex}{D-Brief}& ND & 46.2 & -- & 51.3 \\[-0.9ex]
        \raisebox{1.1ex}{(4 bytes)}& HD & 53.3 & 43.9 & -- \\

        \\
        & LY & 36.2 & 39.9 & 64.9 \\
        mcRBM & ND & 46.2 & 34.5 & 56.1 \\
        (8 bytes) & YM & 43.4 & 37.4 & 53.0 \\
        & LY/ND/HD & 40.5 & 36.6 & 55.4 \\
        \bottomrule
    \end{tabular}}
\caption{Error rates, i.e.\ the percent of incorrect matches when 95\% of the true matches are found. All numbers for GRBM, spGRBM and mcRBMs are given within $\pm 0.5\%$. Every subtable, indicated by an entry in the \emph{Method} column, denotes a descriptor algorithm. Descriptor algorithms that do not require learning (denoted by -- in the column \emph{Training set}) are represented by one line. The numbers in the columns labeled LY, ND and HD are the error rates of a method on the respective test set for this scene. Supervised algorithms are not evaluated (denoted by --) on the scene they are trained on. The \emph{Training set} LY/ND/HD encompasses 1.2 million patches of all three scenes; this setting is only possible for unsupervised learning methods. (a) Error rates for several unsupervised algorithms without restricting the size of the learned representation. GRBM, spGRBM and mcRBM learn descriptors of dimensionality 512. $(L_1\ell_1)$ denotes that the error rates for a method are with respect to $\ell_1$ normalization of the descriptor under the $L_1$ distance. (b) Results for compact descriptors. BRIEF (32 bytes) \cite{calonder2012a} and BRISK (64 bytes) \cite{leutenegger2011a} are binary descriptors, SURF \cite{bay2006a} is a real valued descriptor with 64 dimensions. BinBoost \cite{trzcinski2012c}, ITQ-SIFT \cite{gong2012a} and D-Brief \cite{trzcinski2012b} learn compact binary descriptors with supervision. Numbers for BRIEF, BRISK, SURF, BinBoost and ITQ-SIFT are from \cite{trzcinski2012c}.}\label{tab:res}
\end{table}

Table~\ref{tab:res} consists of two subtables. Table~\ref{tab:res}a presents the error rates for GRBM, spGRBM and mcRBM when no limitations on the size of representations are placed. Table~\ref{tab:res}b only considers descriptors that have an overall small memory footprint. For GRBM and spGRBM we use the activations of the hidden units given a preprocessed input patch $\Vec{v}$ as descriptor $D(\Vec{v})$ (see eq.~\ref{eq:rbmhid}, section~\ref{sec:grbm}):
\[
D(\Vec{v}) = \sigma(\Vec{v}^T\Vec{\Lambda}^{\frac{1}{2}}\Vec{W} + \Vec{b})
\ 
\] 
For the mcRBM a descriptor is formed by using the activations of the \emph{latent covariance units alone}, see eq.~\ref{eq:covrbmhid}, section~\ref{sec:app:mcrbm}:
\[
D(\Vec{v}) = \sigma(\Vec{P}^T(\Vec{C}^T\Vec{v})^2 + \Vec{c})
\]
This is in accordance with manually designed descriptors. Many of these rely on distributions (i.e.\ histograms) of intensity gradients or edge directions \cite{lowe2004a, mikolajczyk2005a, bay2006a}, structural information which is encoded by the covariance units (see also \cite[Section 2]{ranzato2010c})\footnote{Extending the descriptor with mean units degrades results.}.

\subsection{Distance metrics}
As we explicitly refrain from learning a suitable (with respect to the correspondence task) distance metric with a supervised approach, we have to resort to standard distance measures. The Euclidean distance is widely used when comparing image descriptors. Yet, considering the generative nature of our models we follow the general argumentation of \cite{jia2011a} and choose the Manhattan distance, denoted in this text by $L_1$. We also consider two normalization schemes for patch representations, $\ell_1$ and $\ell_2$ (i.e.\ after a feature vector $\Vec{x}$ is computed, its length is normalized such that $\|\Vec{x}\|_1 =1$ or $\|\Vec{x}\|_2 =1$).

Given a visible input both (sp)GRBM and mcRBM compute features that resemble parameters of (conditionally) independent Bernoulli random variables. Therefore we consider the Jensen-Shannon divergence (JSD) \cite{lin1991a} as an alternative similarity measure. Finally, for binary descriptors, we use the Hamming distance.

\subsection{SIFT Baseline}
SIFT \cite{lowe2004a} (both as interest point detector and descriptor) was a landmark for image feature matching. Because of its good performance it is one of the most important basic ingredients for many different kinds of Computer Vision algorithms. It serves as a baseline for evaluating our models.
We use vlfeat \cite{vedaldi2010a} to compute the SIFT descriptors.

The performance of the SIFT descriptor, $\ell_1$-normalized, is reported (using $L_1$ distance) in Table~\ref{tab:res}a, first entry. $\ell_1$ normalization provides better results than $\ell_2$ normalization or no normalization at all. SIFT performs descriptor sampling at a certain scale relative to the Difference of Gaussians peak. In order to achieve good results, it is essential to optimize this scale parameter \cite[Figure 6]{brown2010a} on every dataset. Table~\ref{tab:res}b is concerned with evaluating compact descriptors: the first entry shows the performance of SIFT when used as a \emph{128-byte} descriptor (i.e.\ no normalization applied, but again optimized for the best scale parameter) with $L_1$ distance.

\subsection{Quantitative analysis}
Table~\ref{tab:res}a shows that SIFT performs better than all three unsupervised methods. \emph{mcRBM}(256, 512/512) performs similar to SIFT when trained on Half Dome, albeit at the cost of a 4.5 times larger descriptor representation. The compact \emph{binary} descriptor (the simple binarization scheme is described below) based on \emph{mcRBM}(64, 576/64) performs remarkably well, comparable or even better than several state-of-the-art descriptors (either manually designed or trained in a supervised manner), see Table~\ref{tab:res}b, last entry. We discuss in more detail several aspects of the results in the following paragraphs.

\paragraph{GRBM and spGRBM} spGRBM performs considerably better than its non-sparse version (see Table~\ref{tab:res}a, second and third entries). This is \emph{not} necessarily expected: Unlike e.g.\ in classification \cite{coates2011b} sparse representations are considered problematic with respect to evaluating distances directly. \emph{Lifetime sparsity} may be after all beneficial in this setting compared to strictly enforced population sparsity. We plan to investigate this issue in more detail in future work by comparing spGRBM to
Cardinality restricted boltzman machines \cite{swersky2012a} on this dataset.

\paragraph{Self-taught paradigm} We would expect that the performance of a model trained on the Liberty dataset and evaluated on the Notre Dame scene (and vice versa) should be noticeably better than the performance of a model trained on Half Dome and evaluated on the two architectural datasets. However, this is not what we observe. In particular for the mcRBM (both architectures) it is the opposite: Training on the natural scene data leads to much better performance than the assumed optimal setting.

\paragraph{Jensen-Shannon Divergence} Both GRBM and spGRBM perform poorly under the Jensen-Shannon divergence similarity (overall error rates are around 60\%), therefore we don't report these numbers in the table. Similar, results for mcRBM under JSD are equally bad. However, if one \emph{scales down} $\Vec{P}$ by a constant (we found the value of 3 appropriate), the results with respect to JSD improve noticeably, see Table~\ref{tab:res}a, the last entry. The performance on the Half Dome dataset is still not good -- the scaling factor should be learned \cite{dahl2010a}, which we also plan for future work. 

\paragraph{Compact binary descriptor} We were not successful in finding a good compact representation with either GRBM or spGRBM\@. Finding compact representations for any kind of input data should be done with multiple layers of nonlinearities \cite{krizhevsky2011a}. But even with only two layers (\emph{mcRBM}$(64, 576/64)$) we learn relatively good compact descriptors. If features are binarized, the representation can be made even more compact (64 bits, i.e. 8 bytes). In order to find a suitable binarization threshold we employ the following simple heuristic: After training on a dataset is finished we histogram \emph{all activations} (values between 0 and 1) \emph{of the training set} and use the \emph{median} of this histogram as the threshold.  

\subsection{Qualitative analysis}
\begin{figure}[t]
\begin{center}
\subfloat[]{\includegraphics{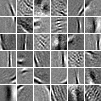}}
\qquad\qquad
\subfloat[]{\includegraphics{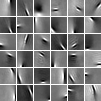}}

\vspace{0.5cm}

\subfloat[]{\includegraphics{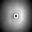}}
\qquad\qquad
\subfloat[]{\includegraphics[scale=1.6]{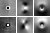}}
\qquad\qquad
\subfloat[]{\includegraphics[scale=0.8]{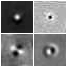}}

\vspace{0.5cm}

\subfloat[]{\includegraphics[scale=0.8]{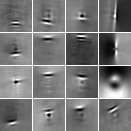}}
\end{center}
\caption{(a) Typical filters learned with spGRBM. (b) Filters from an mcRBM. (c) The pixelwise inverted standard deviations learned with a spGRBM plotted as a 2D image (darker gray intensities resemble lower numerical values). An input patch is elementwise multiplied with this image when computing the latent representation. This figure is generated by training on $32 
\times 32$ patches for better visibility, but the same qualitative results appear with $16 \times 16$ patches. (d) The mcRBM also learns some variants of log-polar filters centered around the DoG keypoint. These are very similar to filters found when optimizing for the correspondence problem in a \emph{supervised} setting. Several of such filters are shown in subfigure (e), taken from \cite[Figure 5]{brown2010a}. Finally (f), the basic keypoint filters are combined with Garbor filters, if these are placed close to the center; the Garbor filters get systematically arranged around the keypoint filters.}\label{fig:filters}
\end{figure}

We briefly comment on the developed filters (Figure~\ref{fig:filters}). Unsurprisingly, spGRBM (Figure~\ref{fig:filters}a) and mcRBM (Figure~\ref{fig:filters}b---these are columns from $\Vec{C}$) learn Gabor like filters. At a closer look we make some interesting observations: Figure~\ref{fig:filters}c shows the diagonal elements of $\Vec{\Lambda}^{\sfrac{1}{2}}$ from a spGRBM\@. When computing a latent representation, the input $\Vec{v}$ is scaled (elementwise) by this matrix, which, visualized as a 2D image, resembles a Gaussian that is dented at the center, the location of the keypoint of every image patch. The mcRBM also builds filters around the keypoint: Figure~\ref{fig:filters}d shows some unusual filters from $\Vec{C}$. They are centered around the keypoint and bear a strong resemblance to discriminative projections (Figure~\ref{fig:filters}e) that are learned in a supervised way on this dataset \cite[Figure 5]{brown2010a}. Qualitatively, the filters in Figure~\ref{fig:filters}d resemble log-polar filters that are used in several state-of-the-art feature designs \cite{mikolajczyk2005a}. The very focused keypoint filters (first column in Figure~\ref{fig:filters}d) are often combined with Gabor filters \emph{placed in the vicinity of the center} -- the Garbor filters appear on their own, if they are too far from the center. If an mcRBM is trained with a fixed topography for $\Vec{P}$, one sees that the Gabor filters get systematically arranged around the keypoint (Figure~\ref{fig:filters}f).

\subsection{Other models}
We also trained several other unsupervised feature learning models: GRBM with nonlinear rectified hidden units\footnote{Our experiments indicate that rmsprop is in this case also beneficial with respect to the final results: It learns models that perform about 2-3\% better than those trained with stochastic gradient descent.} \cite{nair2010b}, various kinds of autoencoders (sparse \cite{coates2011a} and denoising \cite{vincent2008b} autoencoders), K-means \cite{coates2011a} and two layer models (stacked RBMs, autoencoders with two hidden layers, cRBM \cite{ranzato2010a}). None of these models performed as good as the spGRBM.

\section{Conclusion}
\label{sec:conc}
We start this paper suggesting that unsupervised feature learning should be evaluated (i)~without using subsequent supervised algorithms and (ii)~more directly with respect to its capacity to find good low-level image descriptors. A recently introduced dataset for discriminatively learning low-level local image descriptors is then proposed as a suitable benchmark for such an evaluation scheme that complements nicely the existing benchmarks. We demonstrate that an mcRBM learns real-valued and binary descriptors that perform comparably or even better to several state-of-the-art methods on this dataset.

In future work we plan to evaluate deeper architectures \cite{krizhevsky2011a}, combined with sparse convolutional features \cite{kavukcuoglu2010a} on this dataset. Moreover, ongoing work investigates several algorithms \cite{chopra2005a, susskind2011a} for supervised correspondence learning on the presented dataset.
\small{%
\bibliographystyle{abbrv}
\bibliography{./ufldescbib}
}

\normalsize{%
\section{Appendix}
\label{sec:mod}
\subsection{Gaussian-Binary Restricted Boltzmann Machine}\label{sec:grbm}
The Gaussian-Binary Restricted Boltzmann Machine (GRBM) is an extension of the Binary-Binary RBM \cite{freund1994a} that can handle continuous data \cite{hinton2006b, tang2012a}. It is a bipartite Markov Random Field over a set of visible units, $\Vec{v} \in R^{N_v}$, and a set of hidden units, $\Vec{h} \in \{0, 1\}^{N_h}$. Every configuration of units $\Vec{v}$ and units $\Vec{h}$ is associated with an energy $E(\Vec{v}, \Vec{h})$, defined as
\begin{equation}
\label{eq:grbm}
E(\Vec{v}, \Vec{h}; \theta) = \frac{1}{2} \Vec{v}^T\Vec{\Lambda}\Vec{v} -\Vec{v}^T\Vec{\Lambda}\Vec{a} - \Vec{h}^T\Vec{b} - \Vec{v}^T\Vec{\Lambda}\Vec{W}\Vec{h}
\end{equation}
with $\theta = (\Vec{W} \in R^{N_v\times N_h}, \Vec{a} \in R^{N_v}, \Vec{b} \in R^{N_h}, \Vec{\Lambda} \in R^{N_v \times N_v})$, the model parameters. $\Vec{W}$ represents the visible-to-hidden symmetric interaction terms, $\Vec{a}$ and $\Vec{b}$ represent the visible and hidden biases respectively and $\Vec{\Lambda}$ is the precision matrix of $\Vec{v}$, taken to be diagonal. $E(\Vec{v}, \Vec{h})$ induces a probability density function over $\Vec{v}$ and $\Vec{h}$:
\begin{equation}
p(\Vec{v}, \Vec{h}; \theta) = \frac{\exp\bigl(-E(\Vec{v}, \Vec{h}; \theta)\bigr)}{Z(\theta)}
\end{equation}
where $Z(\theta)$ is the normalization partition function,
$Z(\theta) = \int \sum_{\Vec{h}} \exp\bigl(-E(\Vec{v}, \Vec{h}; \theta)\bigr)\,d\Vec{v}$.

Learning the parameters $\theta$ is accomplished by gradient ascent in the log-likelihood of $\theta$ given $N$ i.i.d.\ training samples. The log-probability of one training sample is

\begin{equation}
\log p(\Vec{v}) = -\frac{1}{2}\Vec{v}^T\Vec{\Lambda}\Vec{v} + \Vec{v}^T\Vec{\Lambda}\Vec{a} + \sum_{j}^{N_h} \log\left(1 + \exp\left(\sum_{i}^{N_v}\Vec{v}_i^T(\Vec{\Lambda}^{\frac{1}{2}}\Vec{W})_{ij} + \Vec{b}_j\right)\right)
- Z(\theta)
\end{equation}

Evaluating $Z(\theta)$ is intractable, therefore algorithms like Contrastive Divergence (CD) \cite{hinton2002a} or persistent CD (PCD) \cite{tieleman2008a} are used to compute an approximation of the log-likelihood gradient. The bipartite nature of an (G)RBM is an important aspect when using these algorithms: The visible units are conditionally independent given the hidden units. They are distributed according to a diagonal Gaussian:
\begin{equation}
p(\Vec{v} \mid \Vec{h}) \sim \N(\Vec{\Lambda}^{-\frac{1}{2}}\Vec{W}\Vec{h} + \Vec{a}, \Vec{\Lambda}^{-1})
\end{equation}
Similarly, the hidden units are conditionally independent given the visible units. The conditional distribution can be written compactly as
\begin{equation}
\label{eq:rbmhid}
p(\Vec{h} \mid \Vec{v}) = \sigma(\Vec{v}^T\Vec{\Lambda}^{\frac{1}{2}}\Vec{W} + \Vec{b})
\end{equation}
where $\sigma$ denotes the element-wise logistic sigmoid function, $\sigma(z) = 1/(1+e^{-z})$.
\subsection{Sparse GRBM}\label{sec:spgrbm}
In many tasks it is beneficial to have features that are only rarely active \cite{nair2009a, coates2011b}. Sparse activation of a binary hidden unit can be achieved by specifying a sparsity target $\rho$ and adding an additional penalty term to the log-likelihood objective that encourages the actual probability of unit $j$ of being active, $q_j$, to be close to $\rho$ \cite{nair2009a, hinton2010b}. This penalty is proportional to the negative KL divergence between the hidden unit marginal $q_j = \frac{1}{N} \sum_n p(\Vec{h}_j = 1 \mid \Vec{v}_n)$ and the target sparsity:
\begin{equation}
\lambda_{\mathrm{sp}} \bigl(\rho \log q_j + (1-\rho) \log(1-q_j)\bigr),
\end{equation}
where $\lambda_{\mathrm{sp}}$ represents the strength of the penalty. This term enforces sparsity of feature $j$ \emph{over} the training set, also referred to as \emph{lifetime sparsity}. The hope is that the features for one training sample are then encoded by a sparse vector, corresponding to \emph{population sparsity}. We denote a GRBM with a sparsity penalty $\lambda_{\mathrm{sp}} > 0$ as \emph{spGRBM}.
\subsection{Mean-Covariance Restricted Boltzmann Machine}\label{sec:app:mcrbm}
In order to model pairwise dependencies of visible units gated by hidden units, a third-order RBM can be defined with a weight $w_{ijk}$ for each triplet $v_i, v_j, h_k$. By factorizing and tying these weights, parameters can be reduced to a filter matrix $\Vec{C} \in R^{N_v \times F}$ and a \emph{pooling} matrix $\Vec{P} \in R^{F \times N_h}$. $\Vec{C}$ connects the input to a set of \emph{factors} and $\Vec{P}$ maps factors to hidden variables. The energy function for this \emph{cRBM} \cite{ranzato2010a} is
\begin{equation}
\label{eq:crbmhid}
E_c(\Vec{v}, \Vec{h}_c; \theta) = -(\Vec{v}^T\Vec{C}^T)^2\Vec{P}\Vec{h}_c - \Vec{c}^T\Vec{h}_c
\end{equation}
where $(\cdot)^2$ denotes the element-wise square operation and $\theta = \{\Vec{C}, \Vec{P}, \Vec{c}\}$. Note that $\Vec{P}$ has to be non-positive \cite[Section 5]{ranzato2010a}. The hidden units of the cRBM are still conditionally independent given the visible units, so inference remains simple. Their conditional distribution (given visible state $\Vec{v}$) is
\begin{equation}
\label{eq:covrbmhid}
p(\Vec{h}_c \mid \Vec{v}) = \sigma(\Vec{P}^T(\Vec{C}^T\Vec{v})^2 + \Vec{c})
\end{equation}
The visible units are coupled in a Markov Random Field determined by the setting of the hidden units:
\begin{equation}
\label{eq:gausscrbm}
p(\Vec{v} \mid \Vec{h}_c) \sim \N(\Vec{0}, \Vec{\Sigma})
\end{equation}
with 
\begin{equation}
\label{eq:crbmsigma}
\Vec{\Sigma}^{-1} = \Vec{C}\mathrm{diag}(-\Vec{P}\Vec{h}_c)\Vec{C}^T
\end{equation}
As equation~\ref{eq:gausscrbm} shows, the cRBM can only model Gaussian inputs with zero mean. For general Gaussian-distributed inputs the cRBM and the GRBM can be combined into the \emph{mean-covariance RBM} (mcRBM) by simply adding their respective energy functions:
\begin{equation}
E_{mc}(\Vec{v}, \Vec{h}_m, \Vec{h}_c; \theta, \theta') = E_m(\Vec{v}, \Vec{h}_m; \theta) + E_c(\Vec{v}, \Vec{h}_c, \theta')
\end{equation}
$E_m(\Vec{v}, \Vec{h}_m; \theta)$ denotes the energy function of the GRBM (see eq.~\ref{eq:grbm}) with $\Vec{\Lambda}$ fixed to the identity matrix. The resulting conditional distribution over the visible units, given the two sets of hidden units $\Vec{h}_m$ (\emph{mean} units) and $\Vec{h}_c$ (\emph{covariance} units) is
\begin{equation}
p(\Vec{v} \mid \Vec{h}_m, \Vec{h}_c) \sim \N(\Vec{\Sigma}\Vec{W}\Vec{h}_m, \Vec{\Sigma})
\end{equation}
with $\Vec{\Sigma}$ defined as in eq.~\ref{eq:crbmsigma}. The conditional distributions $p(\Vec{h}_m|\Vec{v})$ and $p(\Vec{h}_c|\Vec{v})$ are still as in eq.~\ref{eq:rbmhid} and eq.~\ref{eq:crbmhid} respectively. The parameters $\theta, \theta'$ can again be learned using approximate Maximum Likelihood Estimation, e.g.\ via CD or PCD\@. These methods require to sample from $p(\Vec{v}|\Vec{h}_m, \Vec{h}_c)$, which involves an expensive matrix inversion (see eq.~\ref{eq:crbmsigma}). Instead, samples are obtained by using Hybrid Monte Carlo (HMC) \cite{neal1993a} on the mcRBM free energy \cite{ranzato2010b}.
}
\end{document}